%% file: jsen.tex
\documentclass[journal,twoside,web]{ieeecolor}
\usepackage{cite}
\usepackage{jsen}
\usepackage{amsmath,amssymb,amsfonts}
\usepackage{algorithmic}
\usepackage{subfigure}
\usepackage{graphicx}
\usepackage{multirow}
\usepackage{textcomp}
\usepackage{wrapfig}
\def\BibTeX{{\rm B\kern-.05em{\sc i\kern-.025em b}\kern-.08em
    T\kern-.1667em\lower.7ex\hbox{E}\kern-.125emX}}
\definecolor{abstractbg}{rgb}{0.89804,0.94510,0.83137}
\begin{document}
\title{Multi-Task and Multi-Modal Learning for RGB Dynamic Gesture Recognition}
\author{Dinghao~Fan, Hengjie~Lu,~Shugong~Xu,~\IEEEmembership{Fellow,~IEEE} and~Shan~Cao,~\IEEEmembership{Member,~IEEE}%

\thanks{}}

\IEEEtitleabstractindextext{%
\fcolorbox{abstractbg}{abstractbg}{%
\begin{minipage}{\textwidth}%
\begin{wrapfigure}[11]{r}{0.4\textwidth}%
\includegraphics[width=0.4\textwidth]{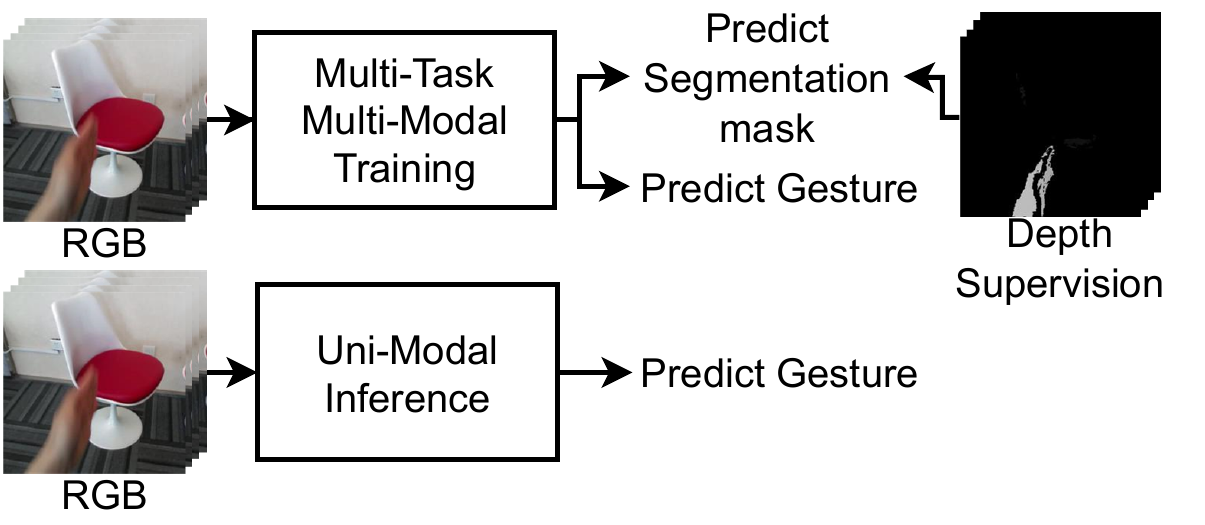}%
\end{wrapfigure}%
\begin{abstract}
Gesture recognition is getting more and more popular due to various application possibilities in human-machine interaction. Existing multi-modal gesture recognition systems take multi-modal data as input to improve accuracy, but such methods require more modality sensors, which will greatly limit their application scenarios. Therefore we propose an end-to-end multi-task learning framework in training 2D convolutional neural networks. The framework can use the depth modality to improve accuracy during training and save costs by using only RGB modality during inference. Our framework is trained to learn a representation for multi-task learning: gesture segmentation and gesture recognition. Depth modality contains the prior information for the location of the gesture. Therefore it can be used as the supervision for gesture segmentation. A plug-and-play module named Multi-Scale-Decoder is designed to realize gesture segmentation, which contains two sub-decoder. It is used in the lower stage and higher stage respectively, and can help the network pay attention to key target areas, ignore irrelevant information, and extract more discriminant features. Additionally, the MSD module and depth modality are only used in the training stage to improve gesture recognition performance. Only RGB modality and network without MSD are required during inference. Experimental results on three public gesture recognition datasets show that our proposed method provides superior performance compared with existing gesture recognition frameworks. Moreover, using the proposed plug-and-play MSD in other 2D CNN-based frameworks also get an excellent accuracy improvement.

\end{abstract}

\end{minipage}}}

\maketitle

\section{Introduction}
\label{sec:introduction}
\IEEEPARstart{R}{ecent} advances in computer vision and pattern recognition have made gesture recognition an accessible and important interaction tool for various applications including human-computer interaction\cite{rautaray2015vision}, sign language recognition\cite{pigou2014sign}, and virtual reality control\cite{lv2015touch}. Most Gesture recognition datasets provide additional data modalities, such as depth frames captured by the depth sensors and 3D skeleton coordinates pre-calculated from depth frames\cite{de2016skeleton}. For multi-modal gesture recognition, multi-stream architecture is proposed to take multi-modality into accounts, such as RGB frames, depth frames, or a precomputed modality like optical flow\cite{miao2017multimodal,zhang2017learning}. These methods have improved performance compared with single-stream architectures. However, multi-modal data inputs require multi-modal sensors, and a multi-stream network requires expensive computation and resource consumption. For real-world applications, RGB cameras are the most common and practical sensors. How to improve the performance of RGB modality with existing multi-modal datasets is a problem to be considered.

\begin{figure*}[htb]
\centering
\includegraphics[width=\textwidth]{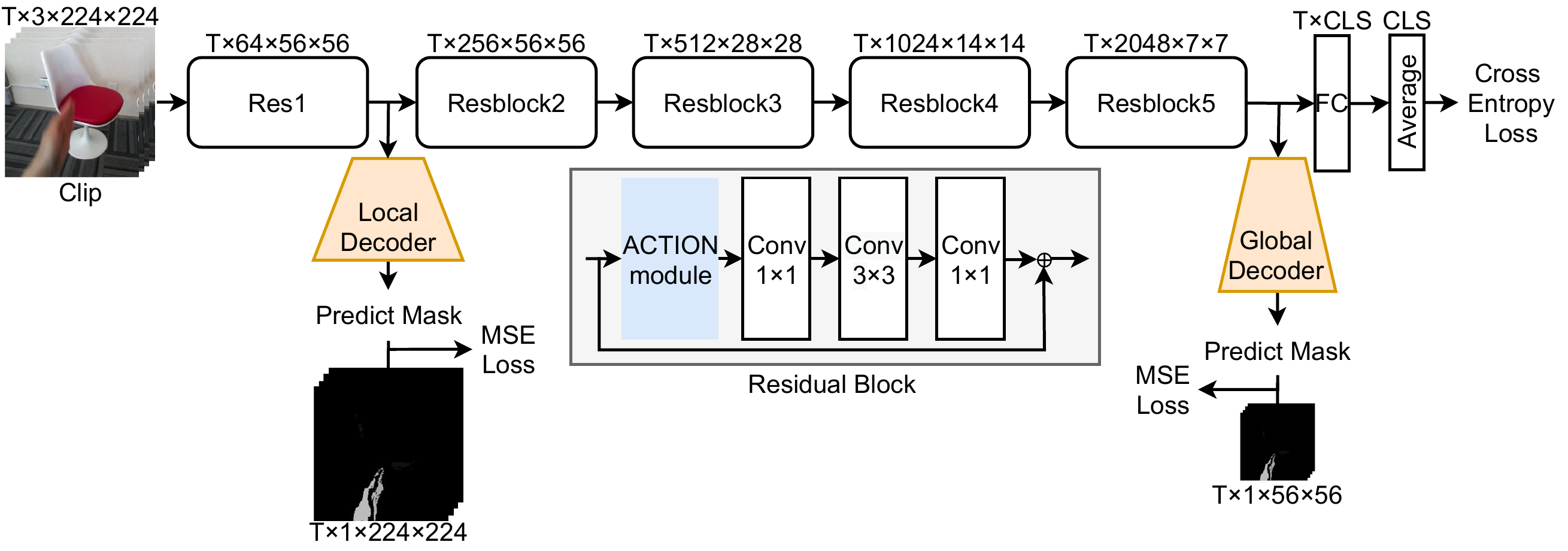}
\centering
\caption{The network architecture is based on the ResNet50 backbone. $CLS$ refers to the number of classes, and $T$ refers to the segment number. The local decoder takes the output of Res1 to predict gesture segmentation, and the global decoder takes the output of Resblock5 as input.}
\label{fig:network}
\end{figure*}

Therefore we propose an end-to-end 2D-CNN with multi-task learning, which can learn gesture segmentation and gesture recognition simultaneously in the training stage to improve accuracy and only require RGB modality during inference. Since depth frames implicitly reflect the position of the gesture, they can be used as supervision of gesture segmentation. The recognition network does not require any modification, the plug-and-play MSD is designed to complete gesture segmentation with different scales in different stages of the network. The MSD module contains two sub-decoder named local decoder and global decoder, which can guide the backbone to segment the gesture and improve the recognition performance. As depth frames can only be considered coarse gesture segmentation maps, we study the binarization of depth frames for generating supervision segmentation maps. We prove that depth frames can be directly used as the ground truth for the supervision of gesture segmentation. The proposed plug-and-play MSD is suitable for other 2D CNN-based recognition frameworks and improves their performances. The framework can automatically ignore background noise in RGB frames and highlight the foreground information. The output segmentation masks in various datasets further demonstrate that the network can distinguish the gesture and background features in the RGB modality. The proposed framework is robust for various datasets regardless of different camera perspectives and different environments, such as the EgoGesture\cite{zhang2018egogesture}, NVGesture\cite{molchanov2016online} and IsoGD dataset\cite{wan2016chalearn}.

To summarize, we propose the following contributions:

$\bullet$ We propose the first end-to-end 2D CNN-based framework with multi-task and multi-modal learning in dynamic gesture recognition. Our framework uses gesture segmentation as a secondary task with depth modality as supervision to improve gesture recognition accuracy.

$\bullet$ A plug-and-play module named MSD is designed in the lower and higher stage of the network, and different scale depth frames are used to supervise the sub-decoder. The MSD module can be removed in inference without additional computation cost.

$\bullet$ Experimental results indicate that our proposed framework presents superior performance than existing gesture recognition frameworks in three widely-used datasets. The accuracy improvement for previous 2D CNN-based frameworks with the proposed plug-and-play module also demonstrates the universality of our method.

\section{Related Work}

\subsection{Sensors for gesture recognition}
Gesture recognition can be practiced mainly by the following four methods: wearable devices\cite{abhishek2016glove,yuan2020hand,wong2021multi}, the 3-dimensional location of hand key points\cite{wen2010intraoperative}, raw visual data, and radar\cite{wang2019ts,skaria2019hand}. The method using glove-based wearable devices requires additional equipment. Even if it provides good results in terms of accuracy and speed, the flexibility of hands is limited, and it is not a natural way of human-computer interaction. Traditional method using hand key points requires an additional step to extract the key points of hands, which brings additional computation time. New method\cite{qi2021multi} uses Leap Motion Controller(LMC), an optical hand tracking module, to get robust and reliable skeletal models. Capturing image sequences containing gestures through image capture sensors, such as RGB cameras, depth sensors, or infrared sensors, is more convenient, practical, and natural because of the intuitiveness of gestures. Recently, the research of gesture recognition on radar has opened a range of new possibilities in intelligent sensing. The radar sensor can instantaneously capture the range and speed of the gesture in each frame signal calculated by fast Fourier transform.

Any gesture recognition system needs to be practical as the purpose is to use it in real-life scenarios. For practical considerations, RGB cameras are still the most common and practical sensors in real life. Therefore the application scenarios of gesture recognition based on RGB cameras are more extensive.

\subsection{3D CNN-based Framework}
3D CNN-based frameworks compute feature maps from both spatial and temporal dimensions, and they have been proven to be effective in spatio-temporal modeling on video classification tasks\cite{tran2015learning,hara2018can,carreira2017quo}. Tran \textit{et al.}\cite{tran2015learning} first experimentally shows 3D convolutional deep networks are good feature learning machines that model appearance and motion simultaneously. I3D\cite{carreira2017quo} inflates the ImageNet\cite{deng2009imagenet} pretrained 2D kernels to 3D kernels for capturing spatio-temporal information. Yu \textit{et al.}\cite{yu2021searching} propose the first neural architecture search (NAS)-based method for RGB-D gesture recognition.

Developments in 3D CNN-based frameworks have significantly improved the performance of dynamic gesture recognition\cite{miao2017multimodal,molchanov2016online,ohn2014hand}. The mainstream of existing methods is based on the structures of 3D-CNN. Although 3D CNN-based approaches have achieved exciting results on several benchmark datasets, the problems for 3D-CNN include overfitting\cite{hara2018can} and slow convergence\cite{tran2018closer}. A large amount of computation inherent in  3D-CNN leads to slow inference, limiting their deployment in practical applications. Recent works such as P3D\cite{qiu2017learning}, R(2+1)D\cite{tran2018closer} have demonstrated that 3D-CNN can be factorized to lessen computations, but the computation is still large when compared to 2D CNN-based frameworks.

\subsection{2D CNN-based Framework}

Although long short-term memory (LSTM)\cite{hochreiter1997long} recurrent neural networks are not 2D CNN-based frameworks, the ability to process sequential data\cite{goodfellow2016deep} is widely used in video recognition. The authors\cite{donahue2015long} apply LSTM for global temporal modeling with the feature extracted by 2D-CNN. However, as the number of video frames increases, the time consumption of LSTM increases proportionally.
 
In \cite{karpathy2014large,simonyan2014two}, video frames are treated as multi-channel inputs to 2D-CNN. TSN\cite{rautaray2015vision} proposes a sparse temporal sampling strategy to learn video representations. Thus 2D CNN-based frameworks are based on a sequence of frames (called segments) sparsely sampled from the entire videos. TSM\cite{lin2019tsm} proposes a zero parameter temporal shift module in 2D ResNet\cite{he2016deep} architecture to exchange information with adjacent frames. There have been various approaches using CNNs to extract spatio-temporal information from video data. Recent works\cite{liu2020teinet,li2020tea,wang2021action} design embedded modules based on the 2D ResNet architecture with motion modeling capabilities.

\subsection{Multi-modal Fusion}

Multi-modal fusion integrates different modalities with different statistical properties. The fusion of information from different modalities is a common approach in CNNs to improve performance. Three main variants for information fusion are proposed. Data-level fusion fuses multiple independent modalities into a single high-dimensional feature vector. Decision-level fusion combines the results from several separate networks trained by different modality data through an algebraic combination rule, including averaging\cite{molchanov2016online,simonyan2014two}, concatenating\cite{zhou2018temporal}, or consensus voting. Feature-level fusion is performed by concatenating feature representations corresponding to multiple modalities. Features from different layers of CNNs can be fused at different layers\cite{feichtenhofer2016convolutional,abavisani2019improving}, or different schemes\cite{miao2017multimodal}.

Our proposed method is related to feature-level fusion. However, different modality feature representations can not be fused explicitly as we design a multi-modal training but unimodal inference framework. We regard depth modality as the supervision for RGB modality input. During inference, multiple modalities are not needed but rather an RGB modality input for classifying. In summary, our proposed multi-task learning framework can make full use of the modality information of the datasets, and uni-modality input during inference is more friendly to real-world applications. Because compared with the previous methods of multi-modal fusion, the number of sensors and deployment costs are reduced.

\section{Proposed Method}

\subsection{Framework}
As described in Figure \ref{fig:network}, our network architecture consists of two main components, a 2D ResNet50 network that generates embeddings for classification and a plug-and-play MSD module that guides the backbone to perceive the location of the gesture. MSD contains two sub-decoder named local decoder and global decoder. Our multi-task learning framework contains one input and three output branches. RGB raw data for the input, classification result, and two segmentation masks for the outputs. We directly use the ACTION module proposed by the latest action recognition framework: ACTION-Net.

The local decoder takes the feature in the early stage as input, upsamples the feature back to the size of the image. The feature in the early stage has high resolution and is more suitable for detailed supervision. The local decoder is set with the same strong constraint as the classification task because the earlier the segmentation of the gesture is obtained, the more helpful to extract the gesture characteristics. The global decoder takes the feature close to the classification head as input. Thus a weak constraint is set for supervising the higher feature. For small resolution in the higher stage, there is no need for obtaining detailed gesture segmentation. So the depth frame size is downsampled for low-resolution supervision. Based on multi-task learning and MSD, the network can distinguish the location of the gesture and significantly reduce the influence of background noise, thereby improving the recognition accuracy.

\begin{figure}[h]
\centering
\subfigure[]{
\includegraphics[width=0.17\textwidth]{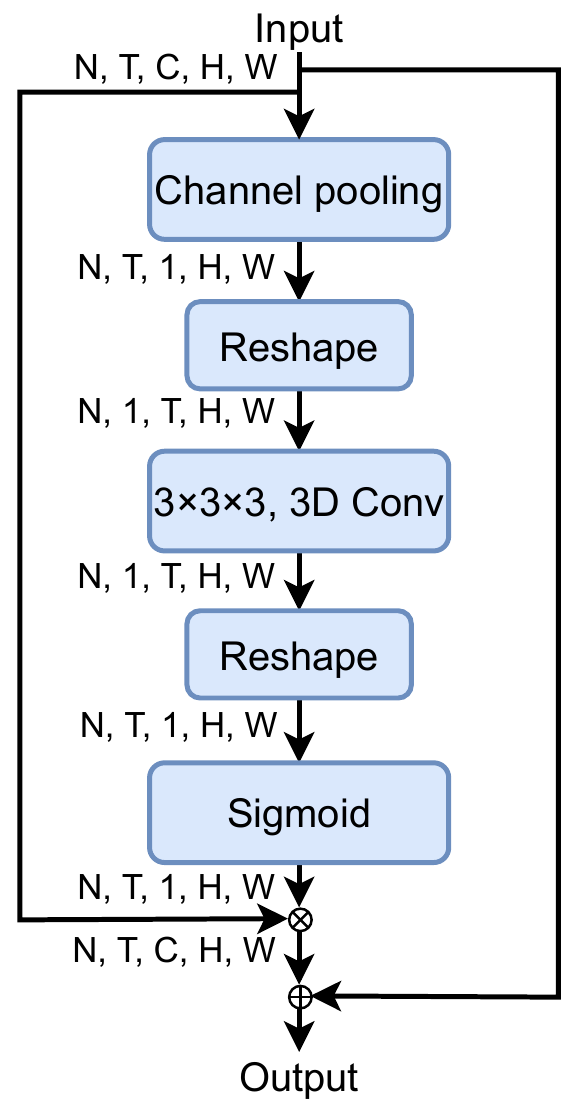}
\label{fig_action_split_a}
}
\subfigure[]{ 
\includegraphics[width=0.17\textwidth]{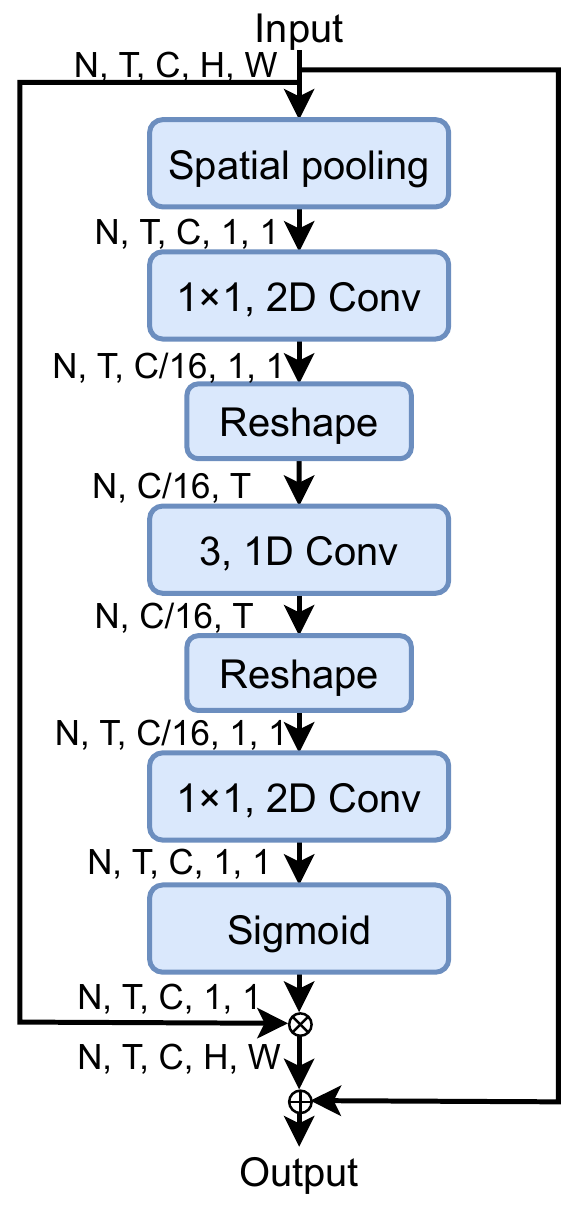}
\label{fig_action_split_b}
}
\subfigure[]{
\includegraphics[width=0.33\textwidth]{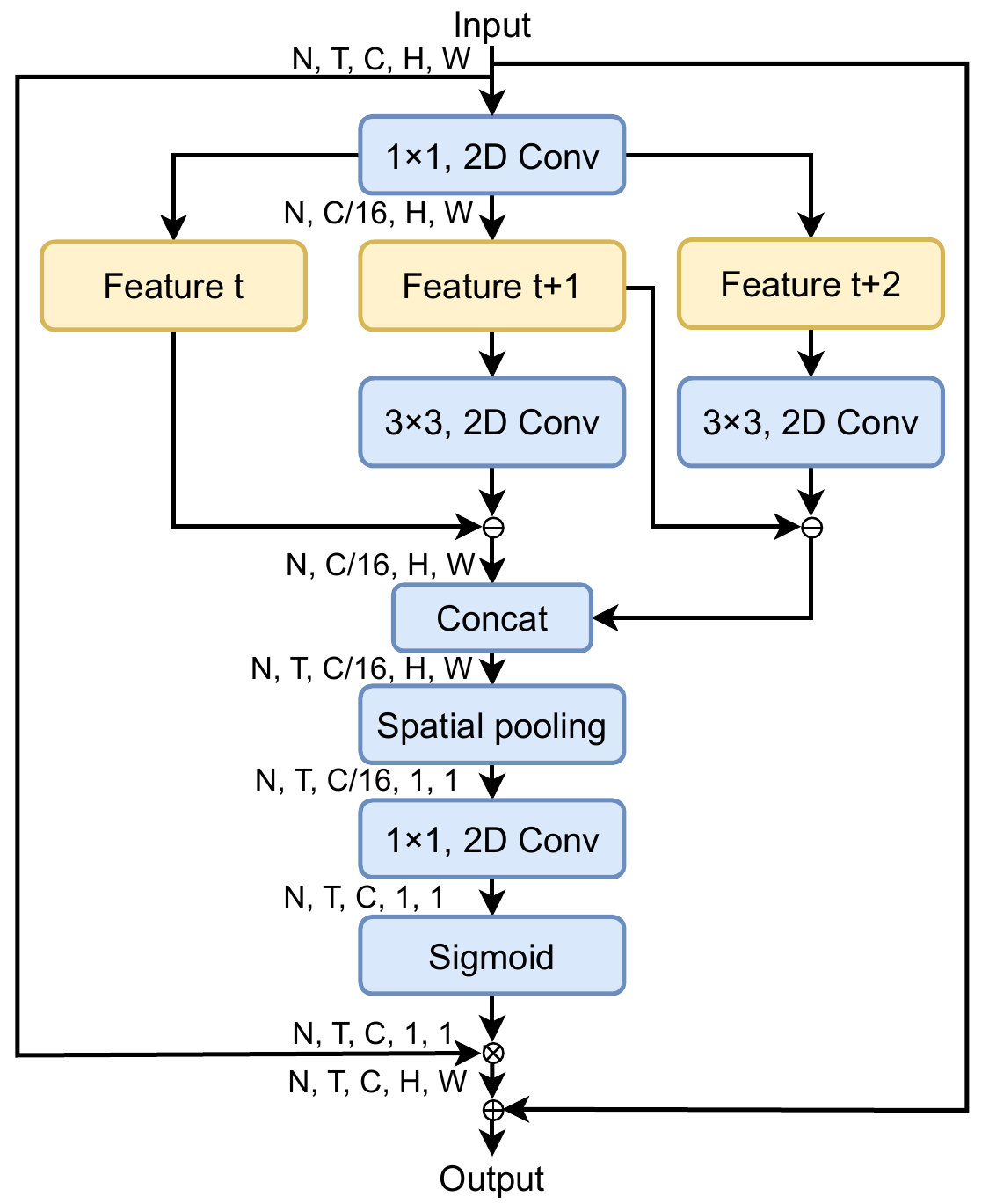}
\label{fig_action_split_c}
}
\caption{Sub-modules in the ACTION module: (a) Spatio-Temporal Excitation, (b) Channel Excitation and (c) Motion Excitation.}
\label{fig_action_split}
\end{figure}

\subsection{ACTION module}

The ACTION module consists of three sub-modules as shown in Figure \ref{fig_action_split}. Before feeding to these sub-modules, the ACTION module uses the temporal shift module proposed by TSM to shift a part of channels on the temporal axis for exchanging information. Notations used in this section are $N$ (batch size), $T$ (number of segments), $C$ (channels), $H$ (height), $W$ (width), and $r$ (channel reduce ratio).

Spatio-Temporal Excitation (STE) is designed for exciting spatio-temporal information, which is described in Figure \ref{fig_action_split_a}. Given an input $\mathbf{X} \in \mathbb{R}^{N \times T \times C \times H \times W}$, a channel pooling is operated for extracting global spatio-temporal feature $\mathbf{F}^{STE}_{pooling} \in \mathbb{R}^{N \times T \times 1 \times H \times W}$. After reshaping  $\mathbf{F}^{STE}_{pooling}$, it is fed to a 3D convolution layer with $3\times3\times3$ kernel size.
 
Channel Excitation (CE) is designed similarly to SE block\cite{hu2018squeeze} as shown in Figure \ref{fig_action_split_b}. CE sub-module learns the importance of different channel characteristics. Given the input $\mathbf{X}$, global spatial information is accessed by spatial average pooling operation, where $\mathbf{F}^{CE}_{pooling} \in \mathbb{R}^{N \times T \times C \times 1 \times 1}$. The number of $r$ is 16. $\mathbf{F}^{CE}_{r} \in \mathbb{R}^{N \times T \times \frac{C}{r} \times 1 \times 1}$ is reshaped and fed to a 1D convolutional layer inserted between two convolutional layers to characterize temporal information. After extracting the temporal feature, a $1\times 1$ 2D convolutional layer is used for upsampling channel dimension back to $C$.

Motion Excitation (ME) has been explored by \cite{jiang2019stm,li2020tea}, which aims to model motion information by adjacent frames effectively as illustrated in Figure \ref{fig_action_split_c}. The same squeeze and unsqueeze strategy is used for downsampling and upsampling the channel dimension, given the feature $\mathbf{F}^{ME}_{r} \in \mathbb{R}^{N \times T \times \frac{C}{r} \times H \times W}$, the main operation can be represented as

\begin{equation}
    \mathbf{F}_{m}=\mathbf{K} *\mathbf{F}^{ME}_{r}[:,t+1,:,:,:]-\mathbf{F}^{ME}_{r}[:, t,:,:,:],
\end{equation}

which is referenced in \cite{jiang2019stm}, where $\mathbf{K}$ is $3\times3$ 2D convolution layer and $\mathbf{F}_{m} \in \mathbb{R}^{N \times 1 \times \frac{C}{r} \times H \times W}$.

The motion feature is then concatenated to each other according to the temporal dimension and 0 is padded to the last element. $\mathbf{F}_{M}=\left[\mathbf{F}_{m}(1), \ldots, \mathbf{F}_{m}(t-1),0\right]$, thus $\mathbf{F}_{M} \in \mathbb{R}^{N \times T \times \frac{C}{r} \times H \times W}$. The $\mathbf{F}_{M}$ is processed by spatial average pooling, the output is $\mathbf{F}^{ME}_{pooling} \in \mathbb{R}^{N \times T \times \frac{C}{r} \times 1 \times 1}$, and upsampling channel dimension operation is the same as the CE sub-module.

For all three sub-modules, a Sigmoid activation is fed in order to get the attention map $\mathbf{M}$. For the STE sub-module, the output size is $\mathbf{F}_{o} \in \mathbb{R}^{N \times T \times 1 \times H \times W}$. For CE and ME sub-modules, the output is $\mathbf{F}_{o} \in \mathbb{R}^{N \times T \times C \times 1 \times 1}$, which can be represented as
\begin{equation}
    \mathbf{M}=\delta\left(\mathbf{F}_{o}\right), 
\end{equation}
where $\delta$ denotes the Sigmoid activation. The final output can be interpreted as
\begin{equation}
    \mathbf{\mathbf{F}^{*}}=\mathbf{X}+\mathbf{X} \odot \mathbf{M}, 
\end{equation}
where $\mathbf{X}$ denotes the output of the temporal shift module, $\mathbf{\mathbf{F}^{*}}$ denotes features $\mathbf{F}^{STE}$, $\mathbf{F}^{CE}$, and $\mathbf{F}^{ME}$. Three separate features are added together and fed to the residual block.

\subsection{Multi-Scale-Decoder}

\begin{figure}[htb]
\centering
\subfigure[]{
\includegraphics[width=0.22\textwidth]{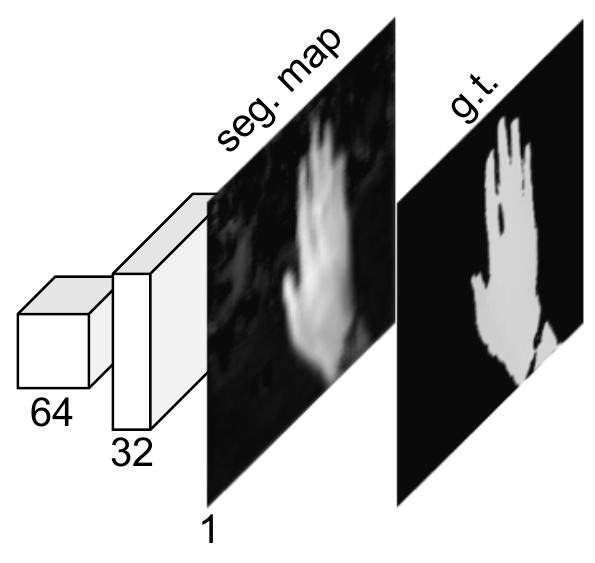}
\label{fig:decoder:a}
}
\subfigure[]{
\includegraphics[width=0.22\textwidth]{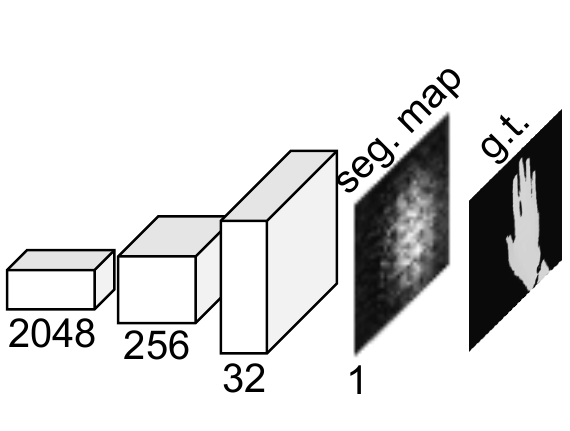}
\label{fig:decoder:b}
}
\caption{The structure of MSD. (a) is the local decoder and (b) is the global decoder, the output is the segmentation mask.}
\label{fig:decoder}
\end{figure}

Works\cite{bambach2015lending,zhu2016two,urooj2018analysis} have already shown that segmented hand images can be a vital feature for recognizing gestures because the network can focus on learning the critical feature. Depth frames significantly reduce the background noise and highlight the foreground gesture. Multi-task learning is designed to perceive depth information. In other words, multi-task learning enables the feature extracted by the network to identify the gesture area well, thereby improving the recognition accuracy. Depth supervised network is composed of MSD, two sub-decoder named local and global decoders are described in Figure \ref{fig:decoder}. We choose features from Res1 and Resblock5 and design two sub-decoder of different scales for the output of different layers. 

The local decoder is designed for the output of Res1. The feature in the early stage has high resolution and is more suitable for detailed supervision. On the other hand, a strong constraint of gesture segmentation in the early stage helps extract gesture recognition information. We set the weight of the MSE loss to be the same as the cross-entropy loss. The feature is upsampled to an image size of $1\times224\times224$ by two transpose convolutional layers. The global decoder is designed for the higher stage, which only requires approximate gesture location. So we use a small scale as supervision for extracting global information. The global decoder is a three-transpose-convolution-layer decoder. Resblock5 feature with $2048\times7\times7$ size is upsampled to a size of $1\times56\times56$. Depth frames are also downsampled to the same size for small-scale supervision by bilinear interpolation. On the other hand, the feature is close to the classification head. A weak constraint is enough for the network to learn gesture location in the higher stage. The weight of the cross-entropy loss and MSE loss for the high stage is $1:0.01$.
\begin{figure*}[h]
\centering
\subfigure[]{
\includegraphics[height=5.5cm]{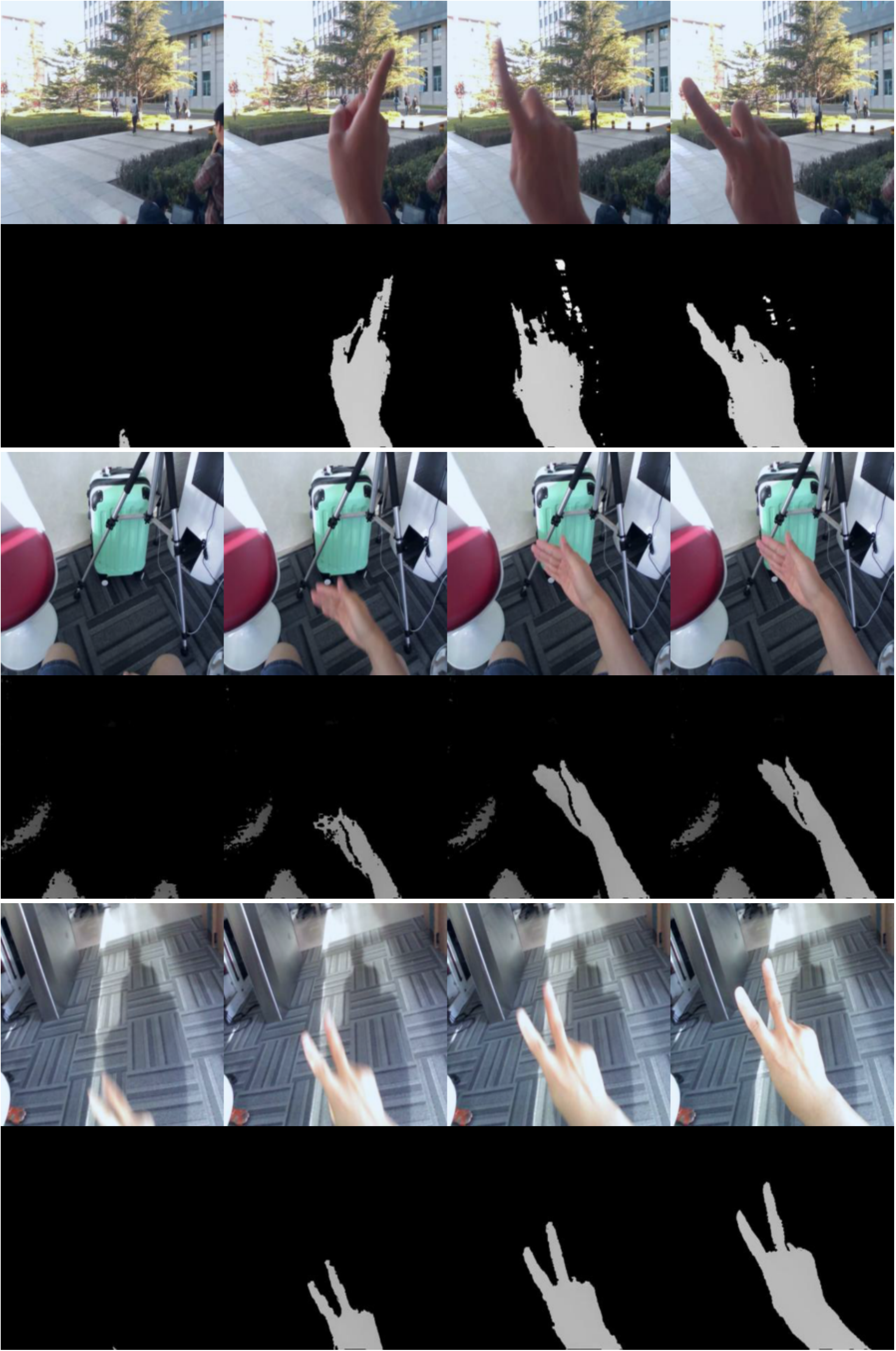}
\label{fig_dataset_a}
}
\subfigure[]{
\includegraphics[height=5.5cm]{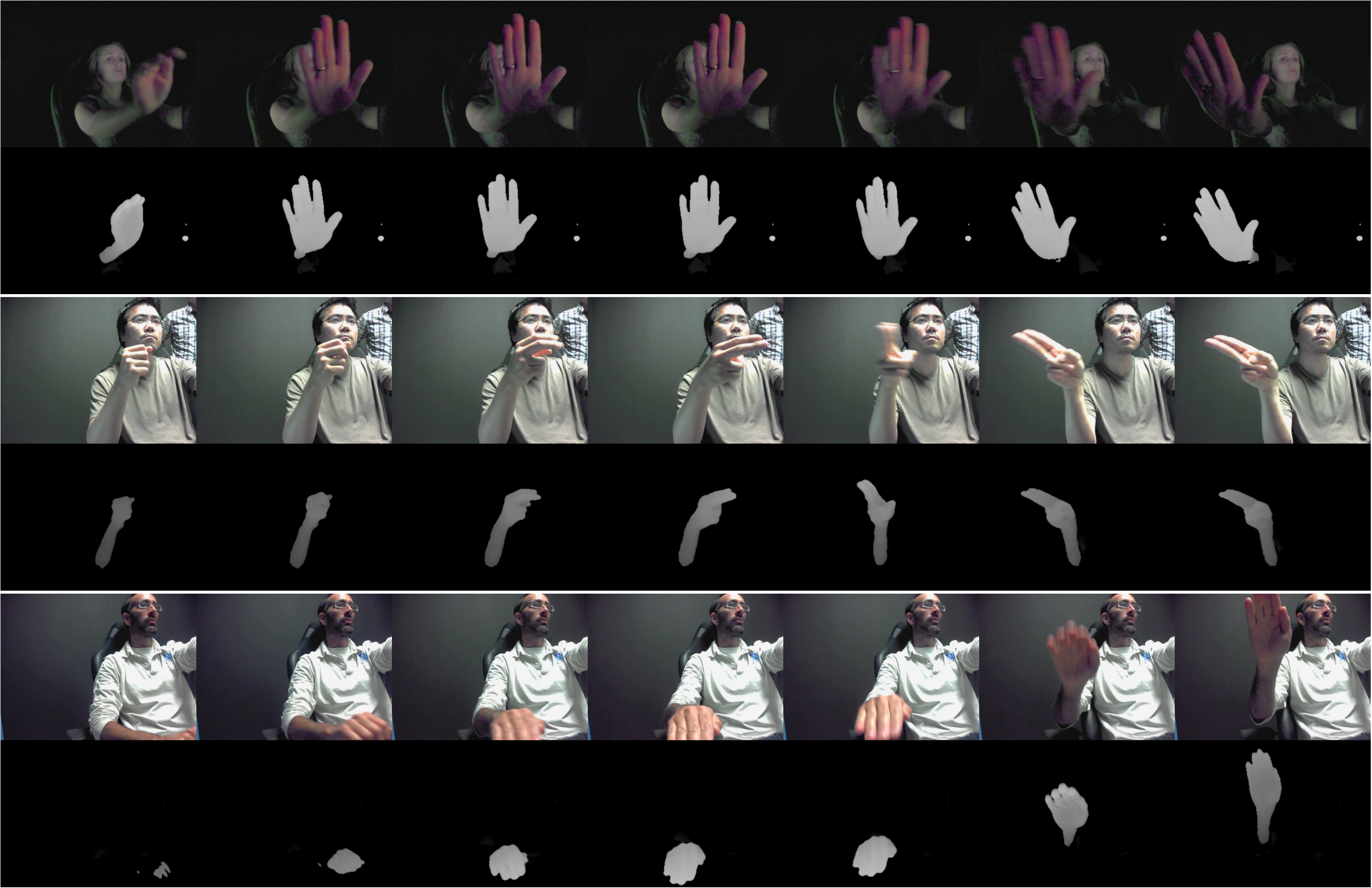}
\label{fig_dataset_b}
}
\subfigure[]{
\includegraphics[height=5.5cm]{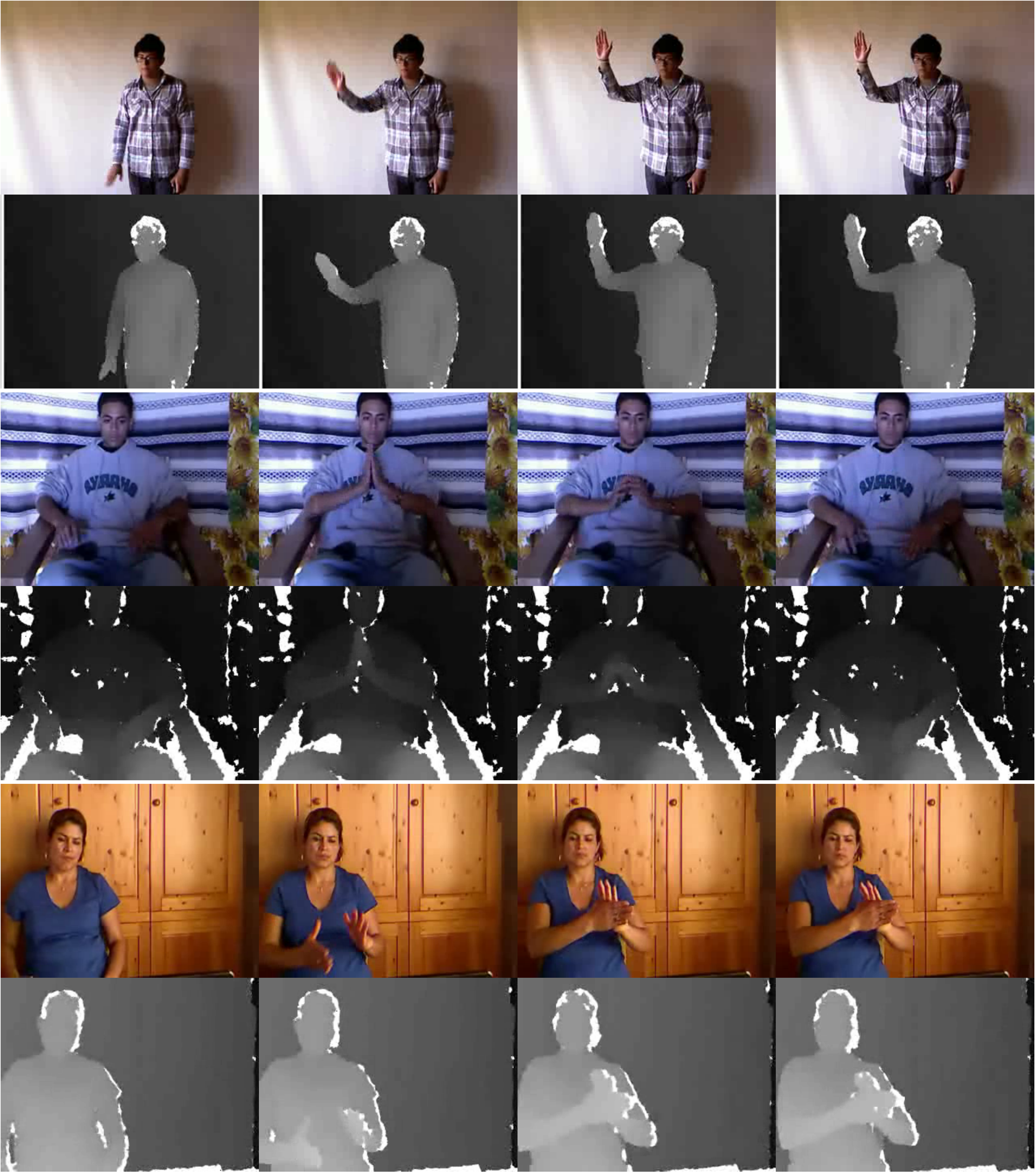}
\label{fig_dataset_c}
}
\caption{RGB and depth frames in each dataset. (a) is the EgoGesture dataset, including indoor and outdoor environments for egocentric hand gesture recognition. (b) is the NVGesture dataset for studying human-computer interfaces, the RGB frames and depth frames are not aligned in the NVGesture dataset. (c) is the IsoGD dataset, we reverse the depth frames to obtain a visualization similar to the previous two datasets.}
\label{fig_dataset}
\end{figure*}

Sample of datasets is shown in Figure \ref{fig_dataset}, such as in Figure \ref{fig_dataset_a} and Figure \ref{fig_dataset_c}, some background in ground truth depth frames is too close to the gesture, which may introduce noise. Segmentation+LSTM\cite{chalasani2019simultaneous} binarizes the depth frames as ground truth. We directly consider depth frames as ground truth as binarizing the depth frames will lose distance information. We conduct experiments for the effect of binarization. The performance in EgoGesture and NVGesture datasets do not improve, which means the network can learn the position of the gesture correctly regardless of the binarization.

Our network can simultaneously supervise the lower feature and higher feature with MSD. It can make the network pay attention to the gesture feature easier. Two sub-decoder complement each other so that the feature extracted by the network can distinguish the gesture area well. Through the supervision by the depth modality, our network can learn the characteristics of the gesture more robustly, avoiding various effects such as environment and lighting and achieving better performance.

\subsection{Network Architecture}
The main architecture is 2D ResNet50. All tensors outside the ACTION module are 4D, $(N \times T, C, H, W)$. The input 4D tensor is reshaped to 5D tensor $(N, T, C, H, W)$ before feeding to the ACTION module. The size of RGB frames is fixed to $224\times224$. The network input is a clip with T frames. Thus the input is $T\times3\times224\times224$. The output size of resblock5 is $T\times2048\times7\times7$. After sending to the global average pooling layer, fully connected layer, and temporal average layer, the final output is the probability of classification.

We generate gesture segmentation masks with the same dimensions as the supervised depth frames. The local decoder has a series of 2 transpose convolutional filters with $4\times4$ kernel size, takes in the hidden state, which size is $T\times64\times56\times56$ as input, upsamples the feature back to the size of the frame. The $T\times2048\times7\times7$ size feature is passed through the global decoder, containing three transpose convolutional filters with $4\times4$ kernel size. The global decoder output is a $T\times1\times56\times56$ dimensional segmentation mask.

\subsection{Loss Function}

Our network is a multi-task learning framework. The gesture label is used to supervise the classification of gestures, and the multi-scale depth frames are used to supervise gesture segmentation. We use two terms: a cross-entropy loss and a pixel-wise mean square error (MSE) loss to train the network. For the local decoder, the loss function is as follows:
\begin{equation}
      L_{\ell_{2local}}=\frac{1}{m} \sum_{i=1}^{m}\left(F_{l_i}-\hat{F}_{l_i}\right)^{2}.
\end{equation}
Where $m$ denotes the number of samples, samples here refer to video clips. Symbol $F_{l_i}$ represents the ground-truth, $1\times224\times224$ depth frames, and symbol $\hat{F}_{l_i}$ represents the output feature from the local decoder.   

The loss function is similar in the global decoder:
\begin{equation}
      L_{\ell_{2global}}=\frac{1}{m} \sum_{i=1}^{m}\left(F_{g_i}-\hat{F}_{g_i}\right)^{2}.
\end{equation}
Where symbol $F_{g_i}$ represents the ground-truth, downsampled $1\times56\times56$ depth frames. Symbol $\hat{F}_{g_i}$ represents the output feature from the global decoder.  

Gesture recognition is supervised with cross-entropy loss:
\begin{equation}
    L_{g}=-\frac{1}{m} \sum_{i=1}^{m} \left(y_{i} \log \hat{y}_{i}+\left(1-y_{i}\right) \log. \left(1-\hat{y}_{i}\right)\right).
\end{equation}
Where $y_{i}$ represents the ground truth gesture label, $\hat{y}_{i}$ represents its prediction result.

Overall, the loss function is as follows:
\begin{equation}
    {L_{\text {total}}=\lambda_{cls}L_{cls}+\lambda_{\ell_{2local}}L_{\ell_{2local}}+\lambda_{\ell_{2global}} L_{\ell_{2global}}}.
\end{equation}
Where $\lambda_{cls}$, $\lambda_{\ell_{2local}}$, $\lambda_{\ell_{2global}}$ are the weights for cross-entropy loss, $\ell_{2}$ loss of local feature and $\ell_{2}$ loss of global feature.

We set the ratio of $\lambda_{cls}:\lambda_{\ell_{2local}}:\lambda_{\ell_{2global}}=1:1:0.01$ because the feature extraction network needs to pay close attention to the gesture texture feature in the early stage, we make a strong constraint on the lower feature. We supervise the feature with a weak constraint for the higher feature to not hinder gesture recognition learning. These settings are evaluated in the ablation study.

\section{Experiments}

During the experiments, we discuss the dataset and training strategy for the network, then compare the proposed method against recent state-of-the-art approaches. The recognition results of previous works plugged in our plug-and-play MSD module are presented to evaluate the performance further.

\subsection{Datasets}

EgoGesture: The EgoGesture dataset is a large-scale egocentric hand gesture recognition dataset designed for VR/AR use cases. It involves 83 classes of gestures in RGB-D modalities from 6 diverse indoor and outdoor scenes. The dataset splits into 14416 training samples, 4768 validation samples, and 4977 testing samples.

NVGesture: The NVIDIA Dynamic Hand Gesture dataset is captured for studying human-computer interfaces. It contains 1532 dynamic hand gestures recorded from 20 subjects inside a car simulator with artificial lighting conditions. This dataset includes 25 classes of hand gestures. The dataset is split by subject into 1050 training videos and 482 test videos.

IsoGD: The Chalearn LAP IsoGD dataset is a large-scale isolated gesture dataset derived from the ChaLearn Gesture Dataset (CGD)\cite{guyon2014chalearn}. IsoGD contains 47933 RGB-D gesture videos divided into 249 kinds of gestures performed by 21 individuals. Participants were given access to 35878 labeled training videos and 5784 labeled validation videos.

\subsection{Implementation Details}
\subsubsection{Training}

Given an input video, we first divide it into T segments of equal duration. One frame is randomly selected from each segment to obtain a clip with T frames. The size of the shorter side of these frames is fixed to 224 and scaled randomly with one of $\left\{1, \frac{7}{8}, \frac{3}{4}, \frac{2}{3}\right\}$ scales. These corner cropping and scale-jittering operations are utilized for both RGB and depth data augmentation. Color Augmentation, including brightness, saturation, contrast, and hue, is randomly utilized for $40\%$ of RGB data. The parameter of color-jittering of brightness, saturation, and contrast is 0.8, and for hue, the parameter is 0.2.

The entire network is trained on two Nvidia P100 GPUs with 32GB memory using the PyTorch framework. We adopted stochastic gradient descent (SGD) as an optimizer with a stochastic momentum of 0.9 and a weight decay of $1\times10^{-5}$. Batch size is set as N = 16 when T = 8. Network weights are initialized using ImageNet pretrained weights for the EgoGesture dataset. For NVGesture and IsoGD datasets, we use the model trained on the EgoGesture dataset as the pretrained model. We start with a learning rate of 0.0025. Detail of the learning rate setting is referred to \cite{goyal2017accurate}, which is halving the learning rate when the batch size is halved. For the EgoGesture dataset, we train the network in training and validation subsets and test in the testing subset. The learning rate is reduced by a factor of 10 at 10, 15, 20 epochs and stopped at 25 epochs. We train the network in training subsets and test in the testing subset for the NVGesture dataset. The learning rate is reduced by 10 at 50, 60, 70 epochs and stopped at 80 epochs. For the IsoGD dataset, the network is trained in training subsets and tested in the validation subset. The learning rate is reduced by 10 at 15, 25, 30 epochs and stopped at 35 epochs.

\subsubsection{Inference}

The proposed MSD is no need for inference, which can be removed for fast inference. We utilize the three-crop strategy following \cite{jiang2019stm} for inference. Firstly, scale the shorter side to 256 for each frame and take three crops of $256\times256$ from scaled frames. For the temporal domain, randomly sample 10 times from the full-length video and compute the softmax scores individually. The final prediction is the averaged softmax scores of all clips.

\subsection{Comparison with SOTA methods} 

Our approach is compared with 2D-CNNs and 3D-CNNs. Two sampling strategies are widely adopted in action recognition to create model inputs. The first one, uniform sampling, often seen in 2D CNN-based models, divides a video into multiple equal-length segments and randomly selects one frame from each segment. The other method used by 3D models, dense sampling, directly takes continuous frames as the input. As referred in \cite{chen2020deep}, I3D and TSM work well with both uniform and dense sampling, so we use the original setting when reproducing the previous works.

\begin{table}[htb]
\caption{Recognition performance comparison on the EgoGesture dataset}
\label{ego_table}
\centering
  \begin{tabular}{|c|c|c|c|}
    \hline
    Methods                              & EgoGesture & GFLOPs &  Parameters \\
    \hline
     C3D\cite{tran2015learning}      &    90.6 & 71.7 & 65.6M  \\ 
    \hline
     MTUT\cite{abavisani2019improving} &   92.5 &  88.2 & 12.7M    \\
    \hline
    NAS1\cite{yu2021searching}                &   93.3 &  - & - \\
    \hline
    3D ResNeXt-101\cite{kopuklu2019real} &    93.7 & 13.9 & 47.7M\\

    \hline        
    \hline
    TSM\cite{lin2019tsm}               &    92.1  & 28.2 & 23.7M \\   
    \hline
    TEA \cite{li2020tea}                &    92.3  & 31.1 &  24.3M \\
    \hline
    ACTION-Net\cite{wang2021action}     &    94.2  & 29.8 & 27.9M \\
    \hline
    Ours                             &    \textbf{94.5} & 29.8 & 27.9M \\
    \hline
  \end{tabular}
\end{table}

We compare our approach with the previous state-of-the-art (SOTA) methods on the EgoGesture, NVGesture, and IsoGD datasets, which is summarized in Table \ref{ego_table}, Table \ref{nvg_table}, and Table \ref{iso_table}. Each table is split into two parts, the top part is the result of 3D CNN-based frameworks, and the following part is the result of 2D CNN-based frameworks. To present the advantages of the learned feature, we directly compare our performance with the reported results of previous 3D CNN-based frameworks. The same experimental setting as ours trains the compared 2D CNN-based frameworks.

\begin{table}[htb]
\caption{Recognition performance comparison on the NVGesture dataset}
\label{nvg_table}
\centering
  \begin{tabular}{|c|c|c|c|}
    \hline
    Methods                             & NVGesture & GFLOPs &  Parameters \\
    \hline
     C3D\cite{tran2015learning}         &  69.3   & 71.7 & 65.6M    \\   
    \hline
    PreRNN\cite{yang2018making}      & 76.5    & 71.7   &   65.6M  \\
    \hline
    3D ResNeXt-101\cite{kopuklu2019real}  &   78.6  & 13.9 & 47.7M \\
    \hline
    MTUT\cite{abavisani2019improving}         & 81.3  &  88.2 & 12.7M  \\
    \hline
    NAS1\cite{yu2021searching}                & 83.6 &  -  &   -  \\
    \hline        
    \hline
    TSM\cite{lin2019tsm}   &  80.2 & 28.2 & 23.7M \\   
    \hline
    TEA\cite{li2020tea}         &   80.5  & 31.1 &  24.3M  \\
    \hline
    ACTION-Net\cite{wang2021action} &   80.7    & 29.8 & 27.9M\\
    \hline
    Ours                             &     \textbf{84.0} & 29.8 & 27.9M  \\
    \hline
  \end{tabular}
\end{table}

\begin{figure*}[htb]
\centering
\subfigure[RGB frames]{
\includegraphics[width=0.21\textwidth]{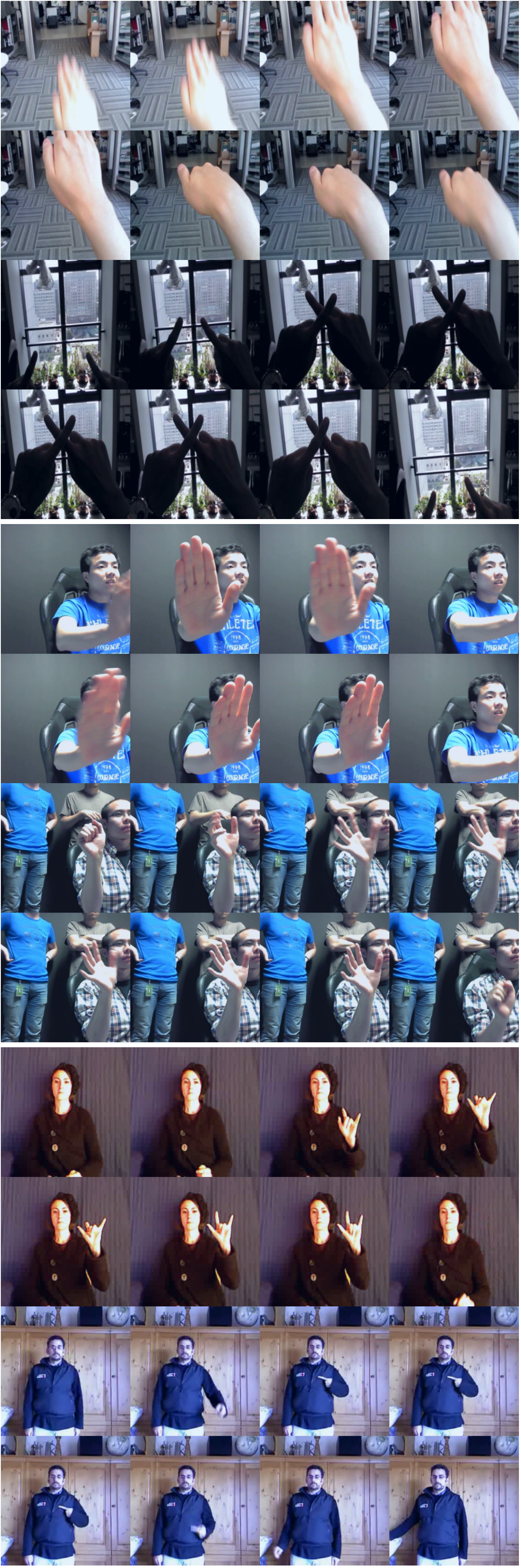}

}
\subfigure[Depth frames]{
\includegraphics[width=0.21\textwidth]{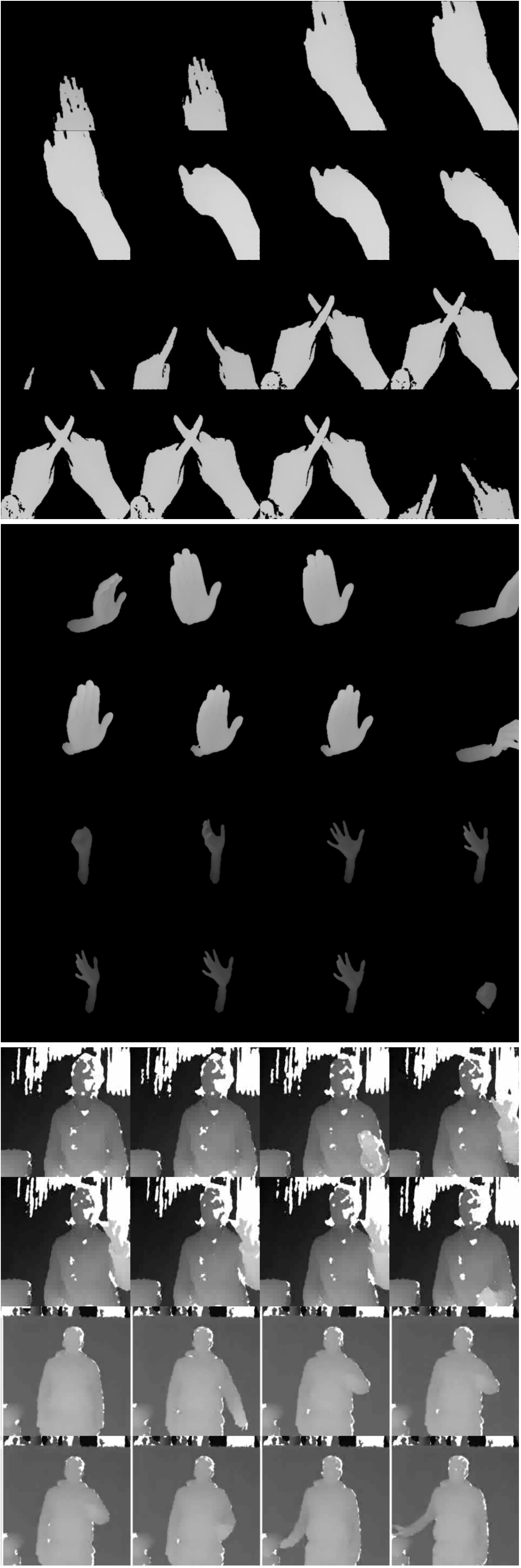}

}
\subfigure[Local decoder predict mask]{
\includegraphics[width=0.21\textwidth]{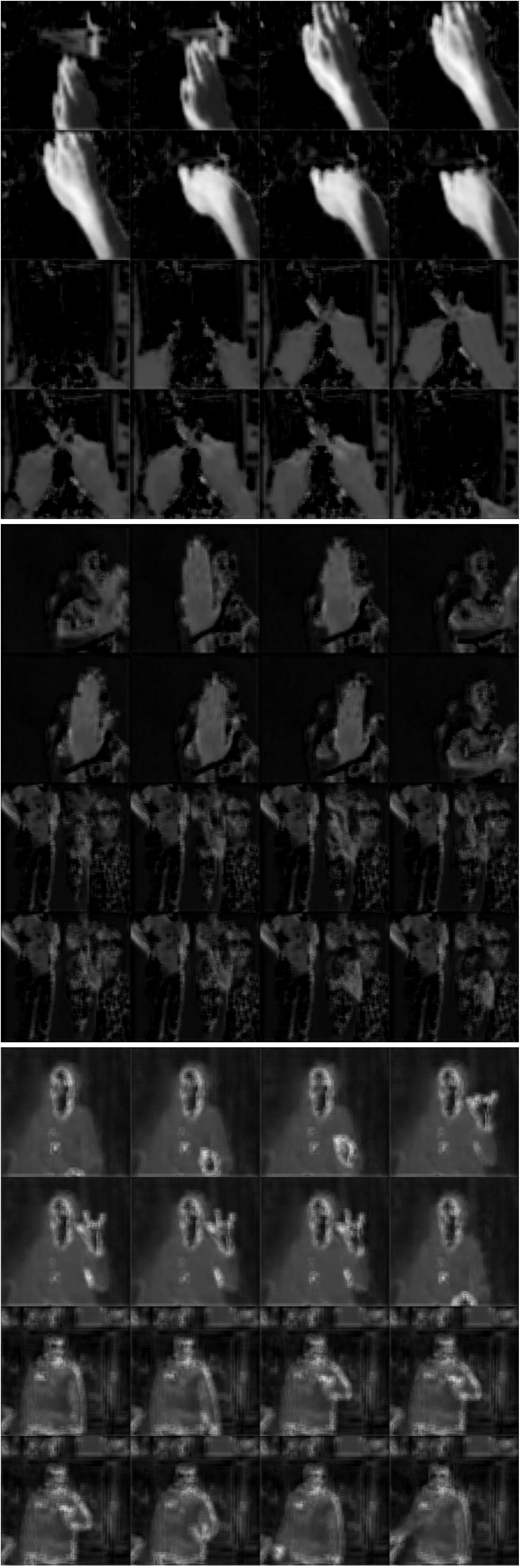}
}
\subfigure[Global decoder predict mask]{
\includegraphics[width=0.21\textwidth]{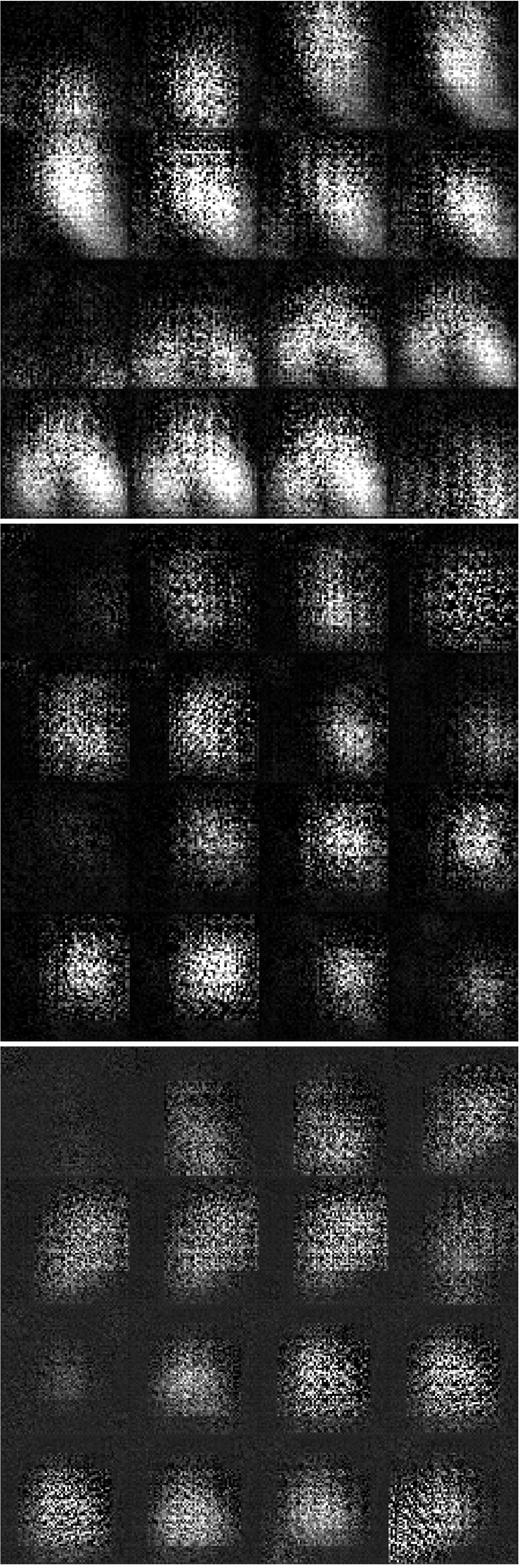}
}
\caption{RGB frames, depth frames, and predict masks generated by local and global decoders in each dataset. The first row is the results in the EgoGesture test subset. The second row is the results in the NVGesture test subset, and the last row is the results of the IsoGD validation subset.}
\label{fig:visible}
\end{figure*}

ACTION-Net is the latest action recognition framework and achieves state-of-the-art performance in 2D CNN-based frameworks. Our framework further improves the accuracy from $94.2\%$ to $94.5\%$ in the EgoGesture dataset. For the NVGesture dataset, the accuracy is increased from $80.7\%$ to $84.0\%$, and $65.0\%$ to $66.9\%$ in the IsoGD dataset. The experimental results demonstrate the effectiveness of our multi-task learning framework. The network has greatly improved performance by using depth frames to learn foreground features. The training strategy only increases 8.5M training parameters and 4.37 GFLOPs. The MSD is removable during inference, which does not increase computation cost. Furthermore, all 3D CNN-based frameworks require more than 32 frames input for better performance. Our method with 8 RGB frames as input achieves superior performance than other methods in three datasets, even the NAS-based method. We do not show the GFLOPs and parameters as the NAS-based framework does not have a fixed model.

\begin{table}[htb]
\caption{Recognition performance comparison on the IsoGD dataset}
\label{iso_table}
\centering
  \begin{tabular}{|c|c|c|c|}
    \hline
    Methods                              & IsoGD  & GFLOPs &  Parameters \\
    \hline
        C3D\cite{tran2015learning}                 & 40.2  & 71.7 & 65.6M \\
    \hline
    ResC3D\cite{miao2017multimodal}      & 45.1 &  77.9 & 38.4M \\
    \hline
    3D ResNeXt-101\cite{kopuklu2019real}   &         52.1  & 13.9 & 47.7M\\
    \hline        
    attentionLSTM\cite{zhang2018attention} &    56.0  & 48.7  & 20.7M \\
    \hline
    Redundancy+AttentionLSTM\cite{zhu2019redundancy} & 57.4 & 48.7  &  20.7M\\
    \hline
    NAS1\cite{yu2021searching}       & 58.9  & -  &  -   \\
    \hline
    \hline
    TSM\cite{lin2019tsm}     &    64.1  & 28.2 & 23.7M\\   
    \hline
    TEA \cite{li2020tea}     & 64.5 & 31.1 &  24.3M \\
    \hline
    ACTION-Net\cite{wang2021action} &  65.0  & 29.8 & 27.9M\\
    \hline
    Ours                             &  \textbf{66.9} & 29.8 & 27.9M\\
    \hline
  \end{tabular}
\end{table}

It should be noted that the RGB and depth modalities are well-aligned in the EgoGesture and IsoGD datasets but not aligned in terms of the NVGesture dataset. Our framework still improves accuracy, which indicates that our method has the general ability. For the IsoGD dataset, the most challenging gesture recognition dataset, our method still performs well with limited input frames compared to previous works. It can be seen that through multi-task learning, the network can learn the information from the depth frames, significantly reduce the interference of background noise, thereby improving the accuracy of RGB modality recognition.

\subsection{Universal of proposed strategy}

\begin{table}[htb]
\caption{Recognition performance of previous works plugged in the MSD module}
\label{second_table}
\centering
  \begin{tabular}{|c|c|c|c|c|c|}
    \hline
    Methods                  & EgoGesture & NVGesture  \\
    \hline
    TSM\cite{lin2019tsm}        &    92.1    &  80.2   \\   
    +MSD   &    93.7    & 81.9     \\
    \hline
    TEA \cite{li2020tea}           & 92.3        & 80.5   \\
    +MSD  &  94.2       &   82.2       \\
    \hline

  \end{tabular}
\end{table}

We conduct our proposed multi-task and multi-modal learning strategy to previous published 2D CNN-based frameworks to evaluate the universal of our strategy. We present the recognition results in Table \ref{second_table}. It is clear that with the multi-scale depth frames supervision, the recognition results of previous 2D CNN-based frameworks such as TSM and TEA make a great improvement. Plugging in the MSD module, the accuracy of TSM improves from $92.1\%$ to $93.7\%$.  The accuracy of TEA improves from $92.3\%$ to $94.2\%$ in the EgoGesture dataset. In the NVGesture dataset, the accuracies improve from $80.2\%$ to $81.9\%$, and $80.5\%$ to $82.2\%$, respectively. This result explicitly proves that our proposed strategy is universal for 2D-CNN gesture recognition. Moreover, the strategy only requires a module named MSD with 8.5M training parameters and 4.37 GFLOPs and does not increase inference computation cost.

\subsection{visualization results}

The visualization of segmentation masks of different datasets is displayed in Figure \ref{fig:visible}, zoom in for better visual quality. Different modality data and segmentation masks in the EgoGesture dataset are shown in the first row. The gesture is greatly highlighted in the segmentation masks from MSD. Even in a backlit environment, the network can robustly learn the segmentation of gestures. For the small-scale dataset NVGesture, it should be noted that the depth frames and the RGB frames are not aligned. The network can still handle the hard case such as crowd background, as shown in the second row. In the third row, Although the depth frames quality of this dataset is not good, containing noise which can affect the network learning. The visualization of segmentation masks can still distinguish the foreground and background. The visualization demonstrates the success of the proposed network. The network can generate gesture noise-reduce frames with the help of depth frames supervision.

\subsection{Discussion with similar works}

We compare our works with similar works: MTUT and Segmentation+LSTM. MTUT first takes RGB and depth modalities as individual input and pretrains two networks. Secondly, the modality network with poor recognition performance learns the knowledge of a better modality network at the semantic feature level during finetune. Segmentation+LSTM first train a encoder-decoder to learn gesture segmentation. The ground truth of segmentation is derived from the depth frame. Secondly, the encoder parameters are frozen, and the embedding from the encoder is sent to the LSTM to learn gesture recognition. Finally, the two systems are cascaded and finetune for better performance.

\begin{table}[htb]
\caption{Detailed comparison of works with similar ideas to ours}
\label{fusion_table}
\centering
  \begin{tabular}{|c|c|c|c|}
    \hline
    Method             &   End-to-End &   Frame   &  EgoGesture \\
    \hline
    MTUT\cite{abavisani2019improving}    &    No      &    64      &     92.5  \\   
    \hline
    \multirow{5}*{Segmentation+LSTM\cite{chalasani2019simultaneous}} &    \multirow{5}*{No}   &     8      &     55.4         \\
                       &                &    16          &     81.4       \\
                       &                 &    24         &      88.4     \\
                       &                &   All          &   93.3     \\
                       &                &   All(in papaer)         &   96.9    \\
    \hline
    ours              &     Yes         &      8     &  94.5        \\
    \hline

  \end{tabular}
\end{table}

Compared with MTUT, firstly, ours directly takes the depth modality as the ground truth of multi-task learning. The idea of MTUT is more similar to knowledge distillation. Secondly, the number of training parameters and the computational complexity increase by multiples with the number of modalities, which requires an enormous training cost. Ours only need to add the plug-and-play MSD module with 8.5M training parameters and 4.37 GFLOPs. Thirdly, MTUT needs to pretrain and finetune two modality networks, ours is an end-to-end training framework, which is friendly for training.

For the result of Segmentation+LSTM, the author claims that the framework uses any length sequence given as input from the dataset. Segmentation+LSTM shares the model and the code for the test. We list the effects of different numbers of input frames using uniform sampling. The result of $96.9\%$ is the result from the paper. Compared with Segmentation+LSTM, firstly, it requires a three-step training stage, ours is an end-to-end training framework. Secondly, Segmentation+LSTM is specifically designed so it cannot be applied to other video classification tasks directly. We propose a plug-and-play module that can directly add to other 2D CNN-based frameworks. Thirdly, Segmentation+LSTM divides the task into two parts: static gesture segmentation and video processing. We introduce multi-scale segmentation maps supervision during gesture segmentation, impose segmentation task constraints in different stages of the network, and combine gesture segmentation and recognition. Finally, the accuracy of Segmentation+LSTM is only $55.4\%$ when using the same 8-frame input as ours. The more input images, the higher the accuracy, which will result in slow inference speed.

\subsection{Ablation Study}
In this section, our framework is verified from the influence of MSD, the supervision scale of MSD, and the effect of segmentation maps generation.

\subsubsection{Evaluation of MSD module}

The MSD module is designed for the gesture segmentation. The ratio of the loss function is critical for network training. We do plenty of experiments on MSD. The local decoder is first taken into account. As shown in Table \ref{ablation_dec_table}, gesture segmentation in the early stage can improve the accuracy in both datasets no matter the ratio setting. We suppose that gesture recognition tasks should focus on the gesture earlier, just like human vision, first localization and then recognition. The result is impressive. The increase in the ratio of MSE loss does not reduce the network's performance but improves the accuracy. If the proportion of MSE loss is reduced to $1:0.1$, although the accuracy is higher than the baseline, it can not compare with setting the ratio to $1:1$, confirming our hypothesis.

\begin{table}[!ht]
\caption{Ablation study of MSD module}

\label{ablation_dec_table}
\centering
  \begin{tabular}{|c|c|c|c|}
    \hline
        method &          cross-entropy:MSE & EgoGesture &NVGesture \\
    \hline
      ACTION-Net\cite{wang2021action}&           -              &   94.2       & 80.7     \\
    \hline  
    \multirow{2}*{+local decoder} &    1:0.1             &    94.3      & 81.1      \\
                    &     1:1             &    94.4      &  82.3      \\
    \hline
    \multirow{3}*{+global decoder}&    1:1               &    94.1      &   80.3     \\
                   &   1:0.1              &    94.2      &   80.5       \\
                   &     1:0.01           &    94.3      &   82.4      \\
    \hline
    +MSD &  1:1 and 1:0.01          &    94.5     &   84.0      \\
    \hline
  \end{tabular}

\end{table}

The global decoder is designed for supervising higher features. The experimental results show that the accuracy gradually increases with the decrease of MSE loss function proportion. This result is in line with our expectation, as the main task of the proposed network is classification, a strong constraint of gesture segmentation for the higher feature is harmful for classification. In contrast, relatively weak regulation can help the network improve classification accuracy without losing classification information. The final setting is fixed to $1:0.01$.

\subsubsection{Evaluation of supervision scale for gesture segmentation task}

The choice of supervision scale for gesture segmentation should be taken into account as well. The experimental result of the scale ablation study is shown in Table \ref{ablation_scale_table}. We change the scale of depth frames to $56\times56$ and apply it to supervise the local decoder. The structure of the local decoder is replaced with two convolutional filters for only downsampling the number of channels. For the EgoGesture dataset, the accuracy is decreased from $94.4\%$ to $93.8\%$. The accuracy is decreased from $80.7\%$ to $77.8\%$  in the NVGesture dataset. The experiment results indicate that the lower feature requires fine-grained supervision to learn the details of gestures in the early stage.

\begin{table}[!ht]
\caption{Ablation study on the scale of depth frame supervision}
\label{ablation_scale_table}
\centering
  \begin{tabular}{|c|c|c|c|}
    \hline
        method &    scale & EgoGesture &   NVGesture \\
    \hline
      ACTION-Net\cite{wang2021action}&           -             &   94.2 & 80.7   \\
    \hline  
    \multirow{2}*{+local decoder} &   $56\times56$      &   93.8   &   77.8   \\
                    & $224\times224$     &    94.4 &   82.3  \\
    \hline
   \multirow{2}*{+global decoder}&     $224\times224$    &   94.0   & 82.4    \\
                      &     $56\times56$    &    94.3  & 82.4   \\
    \hline
  \end{tabular}

\end{table}

For the global decoder, original depth frames with size $1\times224\times224$ are taken to be the supervised information. The initial three transpose convolutional filters are expanded to five for gradually upsampling the feature size. Experimental results show that for the higher feature, large-scale supervision is not required. It not only reduces the accuracy but also increases training computation. The higher stage, which is close to the classification head, no longer needs detailed segmentation results. Global location is more important for the higher feature.

\subsubsection{Evaluation of the segmentation maps generation}

The original purpose of binarizing the depth frames is to remove background noise further. We conduct experiments about the threshold of depth frames binarization. For raw depth frames in EgoGesture, NVGesture, and IsoGD datasets, the threshold is set to 10. If the pixel value is higher than the threshold, the pixel will be set to 255. The result in Table \ref{ablation_bib_table} shows that directly using the depth frames as supervision already achieves the best performance. For EgoGesture and NVGesture datasets, binarizing the depth frames does not increase the accuracy. Because gestures are highlighted correctly, and the background only contains low pixel value noise in these datasets. The network can automatically ignore the background information in the training stage. Thus we directly use the raw depth frames as supervision in these datasets.

\begin{table}[!ht]
\caption{Ablation study of depth frames binarization in EgoGesture, NVGesture and IsoGD datasets}

\label{ablation_bib_table}
\centering
  \begin{tabular}{|c|c|c|c|c|}
    \hline
        method &        threshold & EgoGesture & NVGesture & IsoGD     \\
    \hline
        ACTION-Net\cite{wang2021action}   &       -    &  94.2     &   80.7      &     65.0     \\
    \hline
    \multirow{2}*{ours}   &      -       & 94.5     &   84.0  &   66.9      \\
                             &      10       & 94.5    &   84.0  &  65.4      \\
    \hline
  \end{tabular}

\end{table}

For the IsoGD dataset, the accuracy increases from $65.0\%$ to $65.4\%$ with the threshold of 10. However, the accuracy increases to $66.9\%$ with the supervision of original depth frames. We check the depth frames of the IsoGD dataset in advance. The quality of depth frames is not good, including incorrect highlight background and the interference of foreground objects. Roughly setting the threshold to 10 will make the incorrectly highlighted background noise, and the gesture cannot be separated appropriately. We also directly chose to use the original depth frames as supervision for the IsoGD dataset.

\section{Conclusion}

This paper proposes an efficient end-to-end multi-task and multi-modal learning 2D CNN-based framework for RGB dynamic gesture recognition. The framework is trained to learn a representation for gesture segmentation and gesture recognition. The framework proposes a plug-and-play MSD module to guide multi-scale features to learn segmentation information with depth modality supervision. The MSD can be removed during inference, which does not increase any calculation cost. The experiments indicate that large-scale supervision with a strong constraint for the lower feature and small-scale supervision with a weak constraint for the higher feature is optimal for the framework. Experiments on multiple public gesture recognition datasets, including EgoGesture, NVGesture, and IsoGD, show that our proposed framework with limited input frames outperforms other proposed methods. The visualization of segmentation masks indicates that the proposed framework reduces the background noise with RGB frames input robustly no matter the modality aligned quality and the depth frames quality. Furthermore, our training strategy's accuracy improvement for previous methods demonstrates that we design a general and effective multi-modal learning strategy.

In the future, we will explore multi-task learning on other video understanding tasks (e.g., temporal localization) with more modalities such as skeleton.

\bibliographystyle{IEEEtran}
\input{jsen.bbl}

\end{document}